# Tracing the Path to Grokking: Embeddings, Dropout, and Network Activation

Ahmed Salah[*], David Yevick

*Department of Physics, University of Waterloo, ON N2L 3G1, Canada*

*Corresponding author email address: asalah@uwaterloo.ca

***Abstract–*** Grokking refers to delayed generalization in which the increase in test accuracy of a neural network occurs appreciably after the improvement in training accuracy This paper introduces several practical metrics including variance under dropout, robustness, embedding similarity, and sparsity measures, that can forecast grokking behavior. Specifically, the resilience of neural networks to noise during inference is estimated from a Dropout Robustness Curve (DRC) obtained from the variation of the accuracy with the dropout rate as the model transitions from memorization to generalization. The variance of the test accuracy under stochastic dropout across training checkpoints further exhibits a local maximum during the grokking. Additionally, the percentage of inactive neurons decreases during generalization, while the embeddings tend to a bimodal distribution independent of initialization that correlates with the observed cosine similarity patterns and dataset symmetries. These metrics additionally provide valuable insight into the origin and behaviour of grokking.

## 1. Introduction

Grokking, discovered by Power et al. [1], refers to neural networks for which the increase in the test accuracy is substantially delayed relative to that of the training accuracy. This effect can be interpreted as resulting from the additional accuracy associated with generalization relative to memorization. Liu et al [2] found similar behavior across numerous different model architectures trained on a wide range of datasets, including those involving images, language, and graphs. These and related studies indicate that weight decay and regularization strongly influence the grokking behavior [2, 3] while Yevick et al. [4] demonstrated that grokking can be controlled by modifying the activation function profile.

Liu et al. [5] further conjectured from a study of transformers that generalization is associated with the appearance of structure in the learned input embeddings as evidenced here from the formation of a circular pattern in the plane of the lowest order principal components (PCA). Additionally, Liu et al. and Fan et al. [2, 8], demonstrated that multiplying the initial weights of the neural network by a scale factor can lead to or increase the magnitude of grokking. Progress metrics [9] that track and interpret neural network behavior throughout training also provide insight into the dynamics of generalization. [10], [11], [12] [13,2]. For example, the rapid growth of a L2 metric applied to small sets of neurons during grokking, has been interpreted as a structural phase transition in which a dense, poorly generalizing subnetwork is converted to a sparse, well-generalizing one [14]. Finally, Doshi et al. [15] found that dropout reduces the tendency of a network incorporating RELU activation functions to reduce the number of active nodes and thus its effective dimensionality. As a result, dropout can accelerate generalization, although this also results in a greater variation in accuracy among successive calculations.

This paper proposes several practical metrics for grokking in neural networks in the context of modular addition that can be employed to predict and analyze delayed generalization. One of these quantifies the variance in test accuracy across multiple stochastic forward passes over a fixed batch of test data in the presence of Monte Carlo (MC) dropout [16]. In addition, a novel Dropout Robustness Curve (DRC) quantifies the decreased variability of grokking with respect to random network fluctuations with increased training. Subsequently, the cosine similarity, [6,7] which quantifies the similarity between embeddings in a high-dimensional spaces is found to forecast the rise in test accuracy, providing insight into the origin of grokking within a modular addition framework. The distributions of the embeddings and the neural network weights as well as the dependence of grokking on initialization are also examined and found to possess similar predictive properties as the cosine similarity. For example, the standard deviation of the embedding and weights distributions increases while the corresponding means converge to zero when the training accuracy begins to rise. Similarly, for RELU activation functions, an increased number of inactive neurons can be employed to predict grokking. Accordingly, these results not only provide a cohesive framework for predicting and potentially controlling grokking but also yield additional insight into the origins of grokking behavior.

## 2. Description of model

In the modular additional model employed below, two integer inputs $i\ and\ j$, are first embedded after which the embedding vectors are concatenated and passed to a two-layer MLP consisting of a 256-neuron hidden layer with RELU activation functions, and an output layer with size P=53. Unless otherwise stated, the network is trained on all combinations of inputs with a test/training fraction of 0.5 with an output given by the result of modular addition, $Y = (i + j)\%P$. The network employs cross entropy loss together with the AdamW optimizer, a weight decay of unity, a learning rate=3x10$^{-4}$ and a Xavier Normal initializer in which weights and biases distributed according to $W \sim N(0, \sigma)$ with

$$\sigma = \sqrt{\frac{2}{n_{in} + n_{out}}} \tag{1}$$

Here $n_{in}$ and $n_{out}$ represent the number of inputs and outputs to each neuron. The initial weights are scaled where indicated by a factor α>1.

## 3. Results

### 3.1 Dropout

The variation in test accuracy resulting associated with additional dropout layers can be employed to forecast the onset of grokking as well as the network's sensitivity to structural fluctuations during the transition from memorization to generalization. To illustrate, after a certain number of training steps, the model weights are held constant and 100 stochastic forward passes are performed on the test set. The test accuracy (left axis) and its variance (right axis) are initially negligible, as evident from Fig. 1 which employs a dropout rate of 0.3 while the training accuracy rises to and remains at its maximum value indicating memorization without generalization. At the onset of generalization, the variance in the values of test accuracy variance increases rapidly, indicating an increased sensitivity of the model to small changes in the network. Subsequently, once both the train and test accuracies attain their maximum values, the variance decreases to a near-zero value.

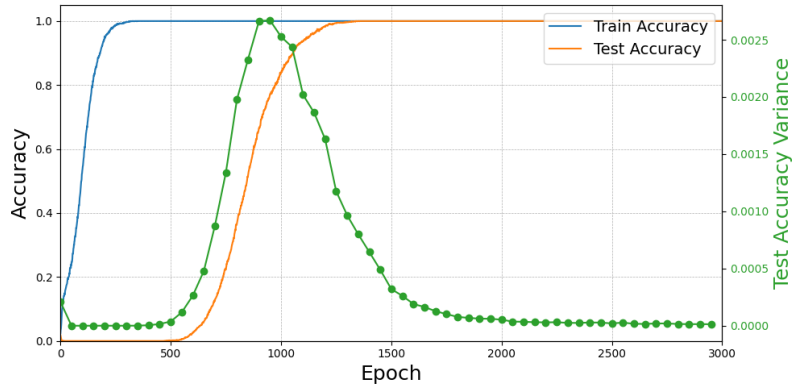

*Figure 1 The training and test accuracies together with the variance in the test accuracy versus epoch number for a dropout fraction of 0.3.*

To characterize further the response of the network to random variations during transition from memorization to generalization a novel dropout robustness curve is presented in Fig. 2. This curve displays test accuracy for dropout rates between 0.0 and 0.9 for the model evaluated at various epochs. During the first $\approx 500$ epochs, the network does not generalize as the test accuracy remains close to zero for all dropout fractions as evident from the orange and cyan lines in Fig.2. After 1000 epochs (green line), the test accuracy starts to increase; however, the test predictions are affected by even small dropout percentages. However, once the model generalizes at $\approx 1500$ epochs (red line), the test accuracy is largely unaffected for dropout rates below 0.2, does not degrade substantially until a dropout rate of 0.5. The purple and brown lines in the figure show the corresponding curves for 2000 and 2500 epochs, respectively, for which the test accuracy has attained its maximum value. This suggests that generalization is accompanied by a decrease in sensitivity to the network structure. This observation could have significant practical implications for predicting grokking in larger networks.

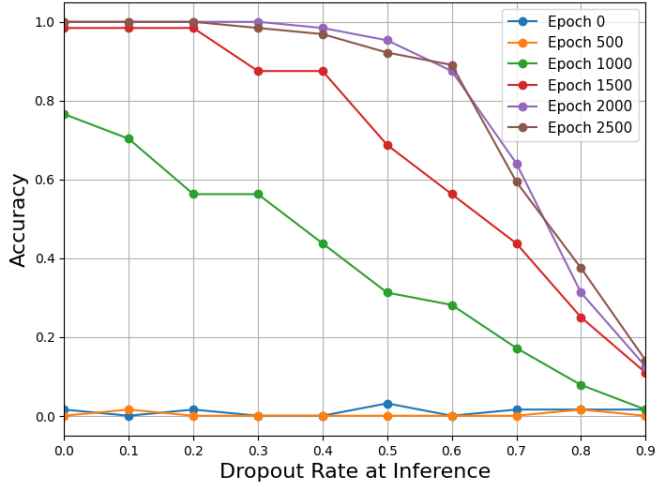

Figure 2 Dropout Robustness Curve

## 3.2 Embedding and representation learning

The variation of the internal structure of the network during training can be analyzed and quantified with the cosine similarity between the embedding vectors. This analysis indicates that grokking is accompanied by the emergence of structured and comprehensible network features. The cosine similarity of two embedding vectors $\boldsymbol{A} = [a_1, a_2, \ldots a_n]$ and $\boldsymbol{B} = [b_1, b_2, \ldots b_n]$ is defined as:

$$C_s(\boldsymbol{A}, \boldsymbol{B}) = \frac{\boldsymbol{A} \cdot \boldsymbol{B}}{\|\boldsymbol{A}\| \, \|\boldsymbol{B}\|} = \frac{\sum_{i=1}^{n} a_i b_i}{\sqrt{\sum_{i=1}^{n} a_i^2} \cdot \sqrt{\sum_{i=1}^{n} b_i^2}} \qquad (2)$$

Averaging over all embedding pairs at each epoch during training yields the result of Fig. 3 where each cell in the heatmap displays the cosine similarity between two embedding vectors with indices given by the x and y values. The values on the diagonal are unity corresponding to the projection of the embedding vector onto itself.

While at initialization, the heatmap of Fig. 1(a) is random, as the training accuracy reaches its maximum value after 300 epochs, codiagonals start to appear in Fig. 1(b). After 450 epochs, bands of codiagonals in Fig. 3 (c) become more discernable, coinciding with the beginning of the growth of the test accuracy and hence the onset of grokking. Once the test accuracy reaches a maximum, the heatmap becomes structured as evident in Fig. 3(d). Since these patterns are not present during memorization and are visible long before grokking occurs, their emergence can be employed to predict grokking. In the case of modular addition, the diagonal bands presumably arise from the periodicity of the modulus. That most of the overlaps are small indicates that the embedding vectors are nearly orthogonal so that the output values can be correctly classified by the network.

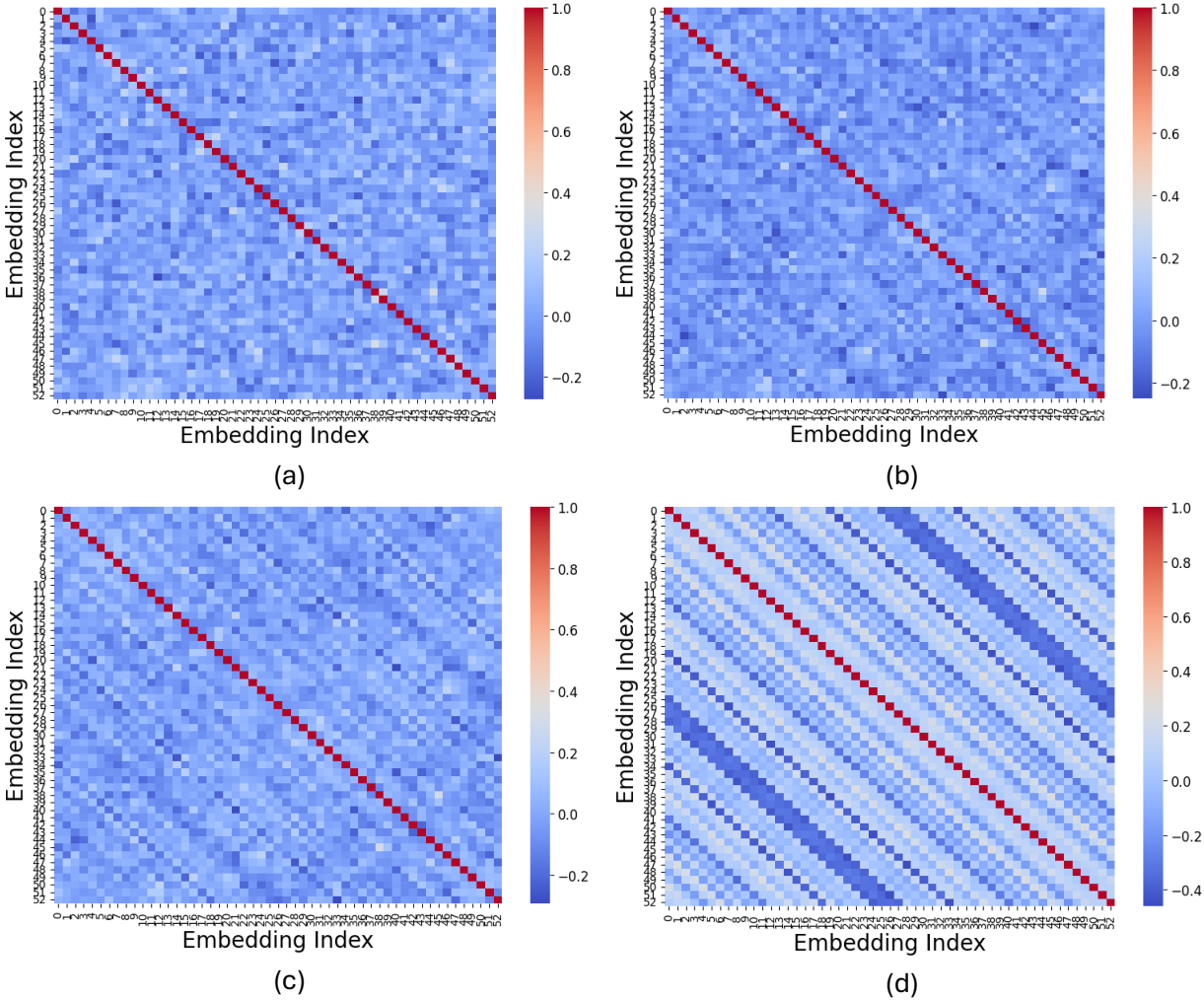

Figure 3 Evolution of cosine similarity of embeddings at epochs (a) zero, (b) 300, (c) 450, (d) 1200.

This evolution of the cosine similarity can also be inferred from the distribution of the embedding values. While initially sampled from a normal distribution, the embedding distribution gradually evolves to two symmetric peaks centered around ±0.4 as shown in Fig. 4(a). Since these peaks emerge concurrently with the formation of the diagonal patterns in Fig. (3); that is, with the rise in test accuracy, this distribution can be alternatively employed to predict the advent of grokking as it quantifies the ability of the network to encode the internal structure of the data.

Indeed, if the weights in the embedding layer are kept constant and not updated in training, the embedding distribution remains a Gaussian function, the cosine similarity heatmap remains unstructured and grokking is suppressed. The magnitude of the embedding peaks decreases with increased weight decay or decreased number of neurons per layer, indicating that the model capacity and regularization influences the grokking behavior. Further, if a ReLU layer is added after the embedding layer, the distribution is shifted to positive values but still remains bimodal with the same peak separation as in the calculation without the layer, however the onset of grokking is significantly delayed. This clearly indicates that the generation of the bimodal structure is intrinsic to learning and not a simple artifact of the activation function.

The distribution of the weights in the layer following the embedding layer retains a normal shape but develops a continuous peak at the origin as the weight decay together with the RELU activation decreases the weight magnitudes as shown in Fig. 4(b). For the last layer, most weights fall in the interval [0.0, 0.2] with a long negative tail as illustrated in Fig. 4(c). In this case, while the weights again decay toward zero, the Relu activation function in the previous layer removes the outputs of many neurons. To compensate and hence enable classification, the final layer weights increase. The extent of this shift, however depends on the choice of cost function and optimization method. However, the magnitude of the positive values is generally smaller than most values in the negative tail of the distribution.

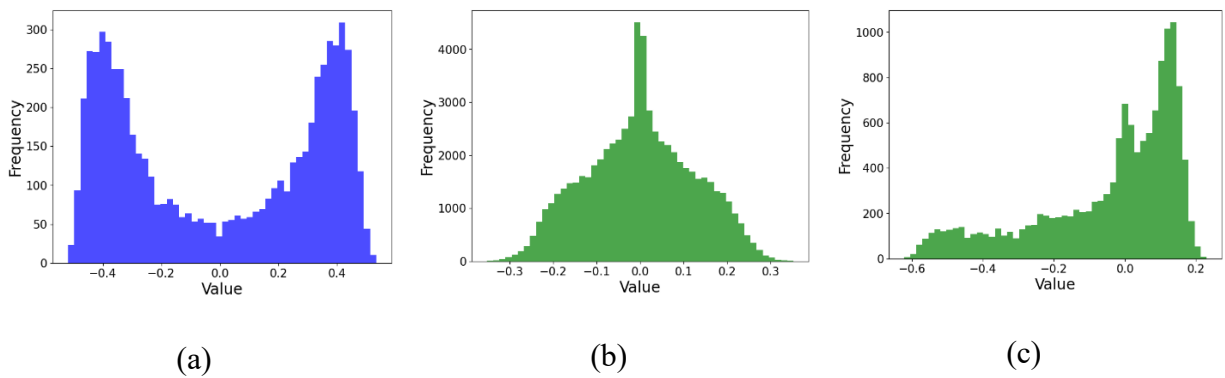

(a)                 (b)                 (c)

*Figure 4 The (a) embeddings, (b)(c) distribution of the weights of the first and second network layers after grokking.*

*3.3 Initialization*

As has been remarked by previous authors, weight initialization has a marked influence on grokking. In this section, this effect will be studied in detail in order to extract practical metrics that enable the prediction and control of grokking. A well-known result is that increasing the initialization amplitudes delays grokking as evident from Fig. 5 (a), which indicates that the delay between the increases in training and test accuracy varies nearly linearly with the scale factor. Presumably, increasing the initialization scale distorts the loss landscape increasing the time for the optimizer to converge to a local minimum. However, the bimodal distribution of the embeddings at ±0.4 is independent of both the scale factor and the initialization profile as it is presumably intrinsic to the symmetries associated with the labeled data. This accordingly indicates that grokking results from these symmetries rather than from overtraining.

After a certain number of epochs, the mean of the weight distribution of the final layer becomes displaced from zero while the standard deviation, similar to that of embedding distribution, increases and then saturates. Hence if the initialization distribution of the layers is initially broad, its width compresses for all layers at the onset of training as is evident from Fig. 6, which refers to (a) the standard unscaled initialization and (b) a widened initial distribution with the scale factor $\alpha = 7$. In this figure, the solid and dashed lines correspond to the standard deviation and mean of the distributions, respectively. That the distributions contract during initial training can also be employed to predict grokking, along the lines of Notsawo et al. [17], who forecast grokking based on oscillations in the loss during the initial epochs.

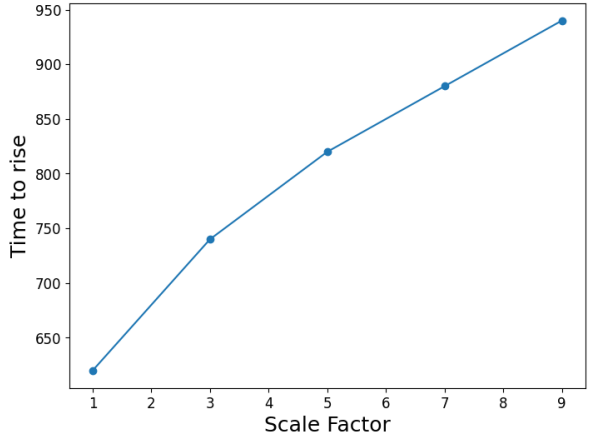
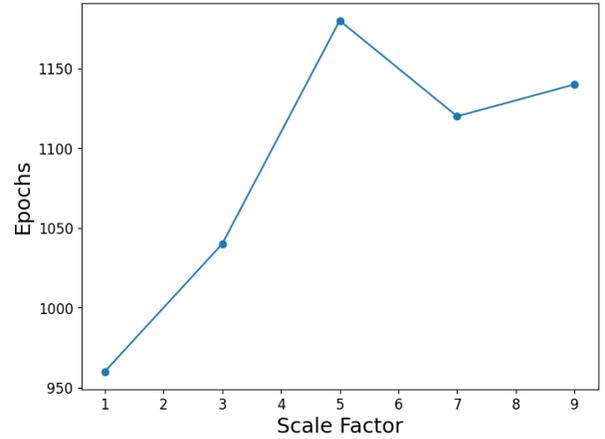

(a)                  (b)

*Figure 5 (a) Epochs before a substantial rise in test accuracy (b) the difference in the number of epochs required for the test and training accuracy to attain near maximum values plotted against the scale factor of the initialization.*

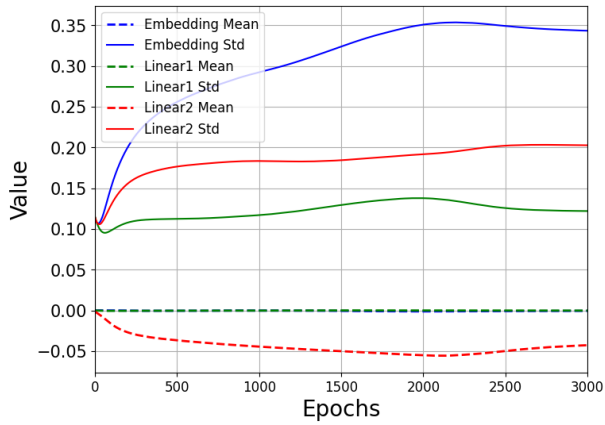
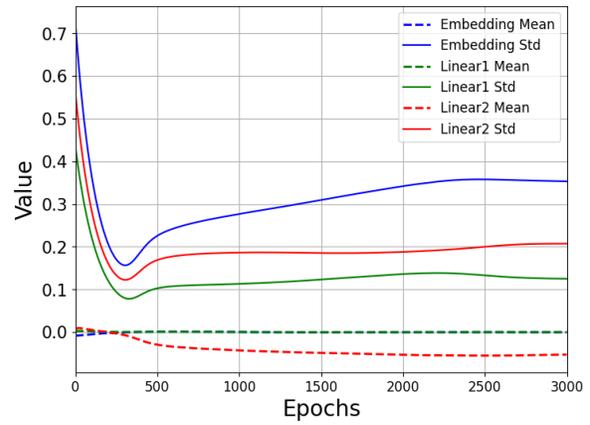

(a)                  (b)

*Figure 6 The variation of embeddings/weights distributions with the number of epochs for (a) unscaled and (b) scaled initialization.*

If all weights are initially positive, grokking still occurs but the rise in the training and test accuracies is delayed. This shift is also evident from the embedding distribution which requires additional epochs to attain a zero mean after which the training accuracy increases. Analogous behaviors are observed for the weight distributions in the other layers. The evolution of embeddings for which the initial distribution is uniformly distributed in the interval [0.4, 0.8] is presented in Fig. 7. In this figure, the distributions not only shift towards zero as evident in Fig.7(b) but also become gaussian, c.f. Fig. 7(c) as the train accuracy starts to rise, after which they widen and subsequently in Fig. 7(d) form two peaks before the test accuracy increases. The mean (dashed lines) and standard deviation (solid lines) of the embeddings/weights distributions when the initialization values are uniformly distributed in the interval [0.4, 0.8] are given in Fig. 8. Compared to Fig. 6(a), this distribution requires additional epochs before the mean decreases to zero while its width decreases followed by a rise in the training accuracy.

Further calculations employing different displacements of the positive initialization interval indicate a nonlinear relationship between initialization bias and generalization. Increasing the displacement reduces the delay between the increase in training and test accuracy, but excessively large shifts damp the rise in test accuracy. The evolution of the test and training accuracies for three different shifted positive uniform distributions together with an unshifted distribution are illustrated in Fig. 9 where the initial weights are uniformly distributed in the intervals [-0.2, 0.2], [0.4, 0.8], [1.2, 1.6], and [1.6, 2] in cases 0 to 3, respectively. In these figures, the solid and dashed lines correspond to the training and test accuracies. Compared to the unshifted distribution shown in blue, cases 1 and 2 which correspond to the orange and green lines respectively, display a smaller delay before the onset of grokking occurs. However the test accuracy for case 4 (red line) with a large initialization shift is significantly reduced and the grokking behavior accordingly becomes anomalous.

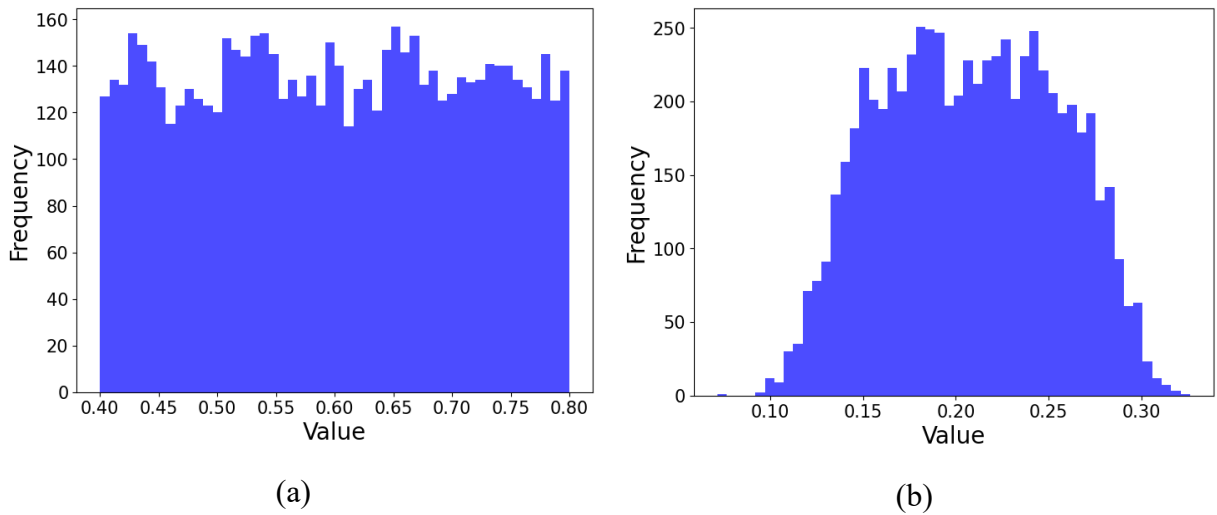

(a)      (b)

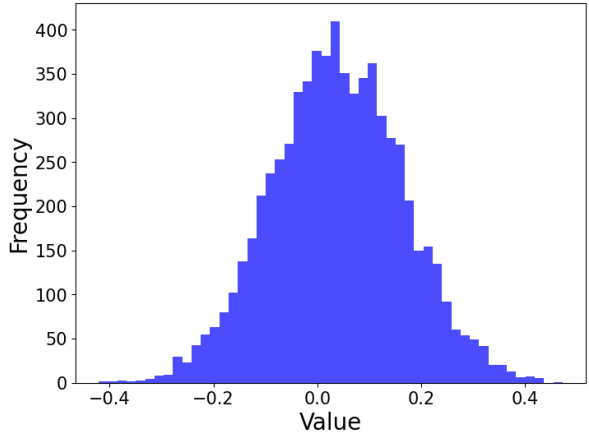 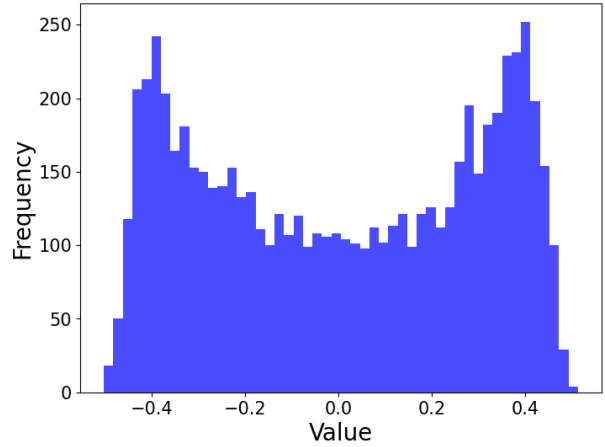

(c)          (d)

*Figure 7 The evolution of the distribution of the embeddings for a shifted uniform within [0.4, 0.8], at epochs (a) zero, (b)150, (c) 600, (d) 1650.*

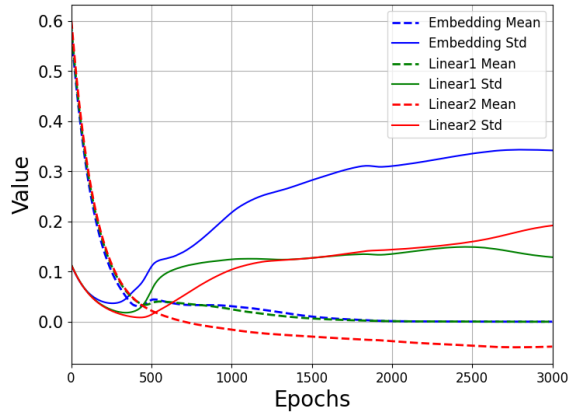

*Figure 8 same as in Figure 5 but for a shifted initialization within [0.4 & 0.8].*

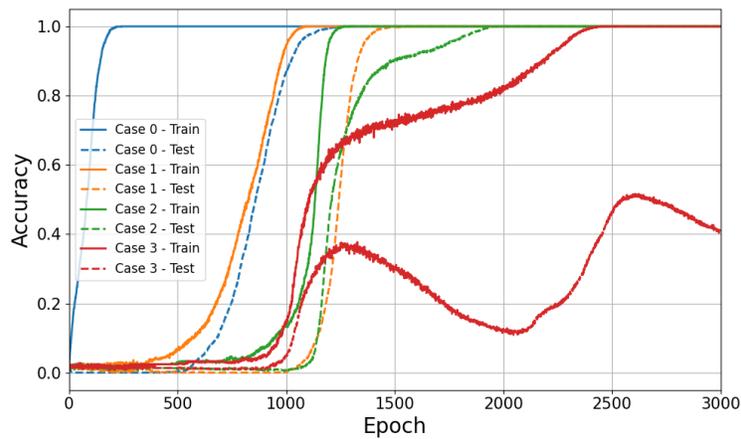

*Figure 9 The training/test accuracy in case of different shifted positive initializations compared to the case of zero centered initialization.*

To isolate further the effects of the initialization, the weight distribution of specific layers was shifted for the embedding layer and then the remaining layers. As shown in Fig. 10, if the distribution in either of these two cases is set to a shifted normal distribution (mean = 0.2, std = 0.1), the resulting delay in grokking is far more pronounced than in the case of an initialization centered at the origin. The test (solid lines) and training (dashed lines) accuracies for zero initialization shift (the blue curve in Fig. 10) reach their maximum values at smaller and larger number of epochs respectively compared to the green curve in the figure for which only the embedding initializations are shifted. This effect is more pronounced if the weight initializations are shifted, as indicated by the cyan curves. Evidently displacing the mean of the initialization of a single layer can significantly affect the number of epochs required to identify patterns in the data. On the other hand, if the embeddings are instead all initialized to the same constant vector while the other layers employ the standard initialization of Section 3.1, learning still occurs since the gradients of the last layers enable optimization. Conversely, initializing the embeddings to a normal distribution while setting the remaining weights to a constant value precludes learning since the embeddings are then not optimized by meaningful gradient values.

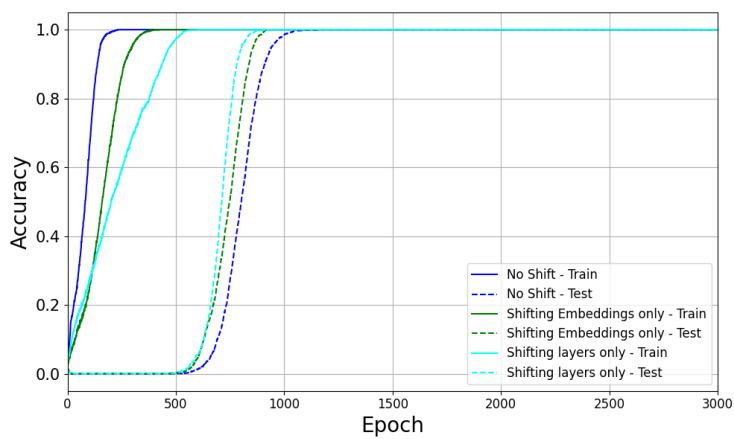

Figure 10 The training/test accuracy if only the initialization of the embeddings is shifted and when only the remaining weights are shifted.

*3.4 Neuron Activations*

The neuron sparsity (the percentage of inactive neurons) during training provides a further metric that can be employed to analyze and potentially forecast grokking behavior. When the training accuracy increases during memorization, the number of inactive neurons falls rapidly.

However, in further epochs the sparsity of the network increases as the optimizer locates symmetries that reduce the effective number of degrees of freedom while the weight decay factor decreases the weight amplitudes. Indeed, the evolution of the sparsity can be controlled through the weight decay parameter as the minimum number of inactive neurons occurs after a smaller number of epochs for larger decay values, resulting in a smaller delay before the onset of grokking. This is consistent with the well-known dependence of the delay on the decay parameter and indicates that the evolution of the sparsity provides a further metric that can be employed to predict grokking.

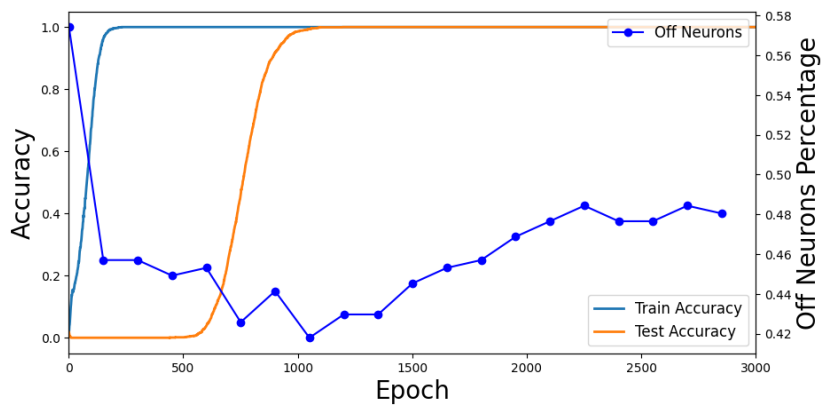

*Figure 11 The training and test accuracy as a function of the percentage of inactive neurons after the Relu unit.*

## 4. Conclusion

From a minimal but easily interpreted model of grokking based on modular arithmetic, several diagnostic metrics that can potentially be employed to predict and quantify grokking have been identified. These include the test accuracy variance under dropout, Dropout Robustness Curves, embedding similarity patterns, weight distributions and neuron activity, all of which correlate with the onset of grokking. In the first case, a rise in the variance of the test accuracy resulting from dropout is a precursor of grokking. This behavior is also manifest in the dropout robustness curve, which characterizes the evolution of the variance in the test accuracy with epoch number and falls rapidly when a network transitions from memorization to generalization. Furthermore, in modular arithmetic, grokking is accompanied by the emergence of symmetric, bimodal distributions of the embedding weights with periodic cosine similarity patterns indicating that the network has adapted to the underlying symmetries of the dataset. The embeddings and weights similarly converge to similar distributions regardless of initialization. Accordingly, tracking statistical properties such as the mean and standard deviation of these distributions provides predictive insight into the grokking behavior of the network. The number of inactive neurons also decreases as the model begins to generalize and accordingly increase the accuracy of its representation of the data. Future work could examine the extent to which these indicators can be extended from a standard benchmark model to realistic problems as well as investigate in detail the relationship between the training hyperparameters and the grokking features.